\documentclass[letterpaper, 10 pt, conference]{ieeeconf}  %

\IEEEoverridecommandlockouts                              %

\usepackage{amsmath}
\usepackage{amssymb}
\usepackage{hyperref}
\usepackage[table]{xcolor}
\usepackage{amsmath}
\usepackage{amssymb}
\usepackage{mathtools}
\usepackage{siunitx}
\usepackage{graphicx}
\usepackage{adjustbox}
\usepackage{float}
\usepackage{wrapfig}
\usepackage{booktabs}
\usepackage{tabularx}
\usepackage{threeparttable}
\usepackage[font=small]{subcaption}
\usepackage[font=small,skip=3pt,tableposition=top]{caption}
\usepackage{microtype}
\usepackage[noadjust]{cite}  %

\usepackage{array}
\newcommand{\PreserveBackslash}[1]{\let\temp=\\#1\let\\=\temp}
\newcolumntype{C}[1]{>{\PreserveBackslash\centering}p{#1}}
\newcolumntype{R}[1]{>{\PreserveBackslash\raggedleft}p{#1}}
\newcolumntype{L}[1]{>{\PreserveBackslash\raggedright}p{#1}}

\newcolumntype{Y}{>{\centering\arraybackslash}X}

\def\figurespace{\vspace{-2.75ex}}

\newcommand{\ours}{AMR}

\title{\LARGE \bf
Aim My Robot: Precision Local Navigation to Any Object 
}
\author{Xiangyun Meng, Xuning Yang, Sanghun Jung, Fabio Ramos\\ Srid Sadhan Jujjavarapu, Sanjoy Paul and Dieter Fox
\thanks{Xiangyun Meng, Sanghun Jung and Dieter Fox are with the Paul G. Allen School of Computer Science \& Engineering, University of Washington, Seattle, WA 98195, USA {\tt\small\{xiangyun,shjung13, fox\}@cs.washington.edu}}
\thanks{Xuning Yang, Fabio Ramos, and Dieter Fox are with NVIDIA, Seattle, WA 98105, USA {\tt\small \{xuningy, ftozetoramos,dieterf\}@nvidia.com }}
\thanks{Sri Sadhan Jujjavarapu, Sanjoy Paul are with Accenture LLP. {\tt\small \{srisadhan.j, sanjoy.paul\}@gmail.com }}
}

\begin{document}

\maketitle

\thispagestyle{empty}
\pagestyle{empty}

\begin{abstract}
Existing navigation systems mostly consider ``success'' when the robot reaches within 1\unit{\meter} radius to a goal. This precision is insufficient for emerging applications where 
the robot needs to be positioned precisely \footnote{We use the word ``precise'' to indicate both low error and high consistency. This is different from the conventional meaning of ``precise'' that only refers to high consistency.} relative to an object for downstream tasks, such as docking, inspection, and manipulation. 
To this end, we design and implement \emph{Aim-My-Robot} (\emph{AMR}), a local navigation system that enables a robot to reach any object in its vicinity at the desired relative pose, with \emph{centimeter-level precision}.
\ours{} achieves high precision and robustness by leveraging multi-modal sensors, precise action prediction, and is trained on large-scale photorealistic data generated in simulation. \ours{} shows strong sim2real transfer and can adapt to different robot kinematics and unseen objects with little to no fine-tuning.\\
Website: \footnotesize{\color{blue}{\url{https://sites.google.com/view/aimmyrobot}}}
\end{abstract}

\section{Introduction}

\begin{figure}[t]
\centering
\includegraphics[width=0.9\columnwidth]{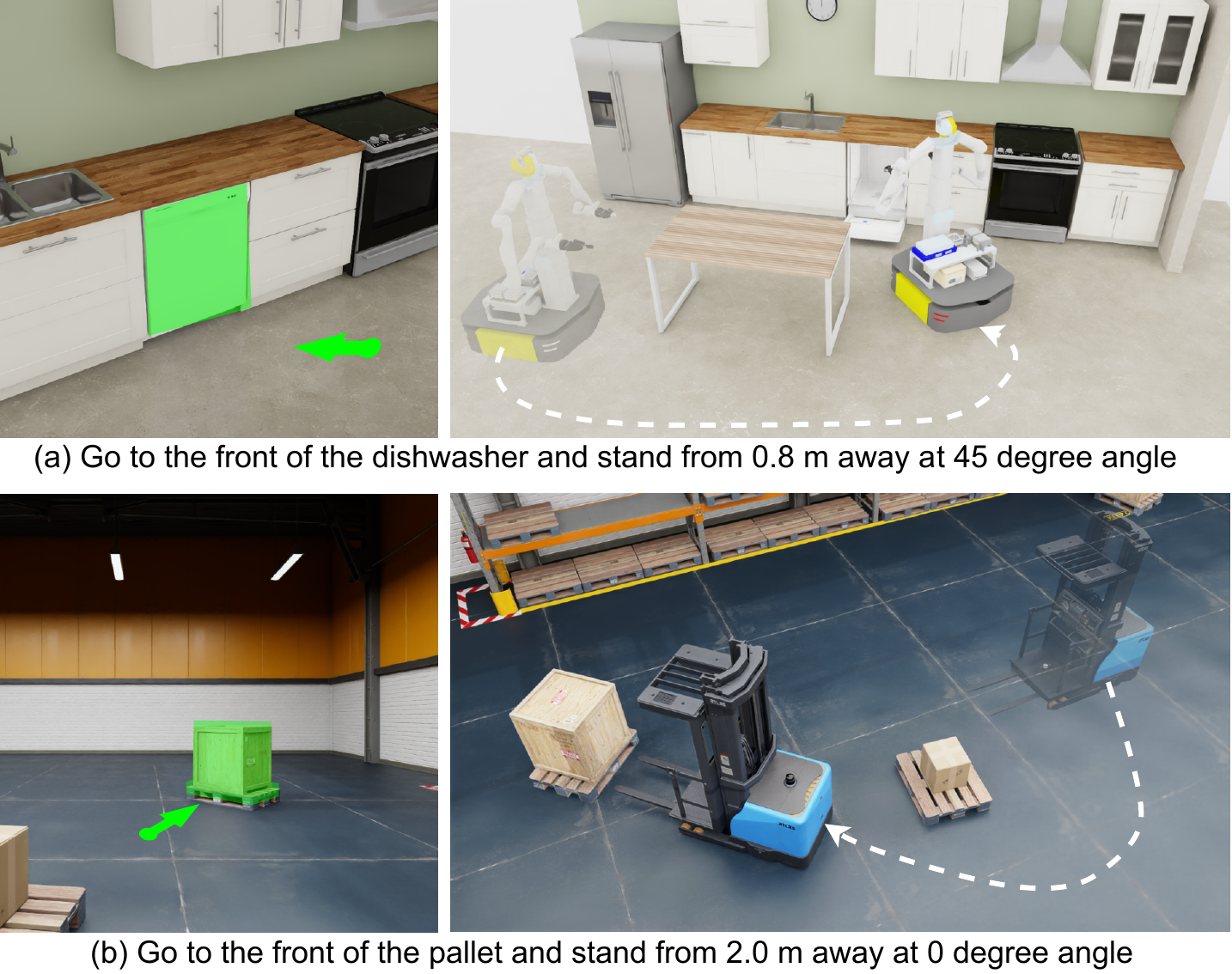}
\caption{\textbf{Overview of \ours{}}. Given a masked image describing the target object and an object-centric pose (relative position and orientation), \ours{} tracks the object while moving, avoids obstacles, and aligns the robot to the target object with centimeter-level precision without maps or object 3D models.}
\label{fig:teaser}
\figurespace
\vspace{-3mm}
\end{figure}

Navigation is a foundational skill that unleashes robots from confined workspaces and lets them interact with the open world. While there has been significant progress in learning-based systems that can navigate without a geometric map~\cite{meng2019scaling,spoc,nomad,saynav,gnm}, understand semantics~\cite{saynav, feature-fields, mopa, chang2023goat, nav-llm}, and follow human instructions~\cite{saynav, chang2023goat, nav-llm,  interactive-nav}, these systems typically consider the goal reached when the robot is within 1\unit{\meter} radius to the goal~\cite{chang2023goat, krantz2022instance, habitatchallenge2023}. This lax definition of success hinders their applicability to the growing need for mobile robots to navigate to objects with precisely. 
For example, in a factory, a forklift must position itself correctly to a pallet so that the fork can be inserted into the pallet without collision (Fig.~\ref{fig:teaser}b); an inspection robot can more clearly read gauges when facing instruments perpendicularly at a proper distance. Similarly, a home robot needs to position itself properly to open a dishwasher (Fig.~\ref{fig:teaser}a), dock to a charging station, or place objects at accurate locations (e.g. \emph{left} side of a table). Lack of precision in the final pose of the robot would incur failures due to collision, out-of-reach, or not adhering to task requirement.

Precision navigation requires a robot to understand the geometry of relevant objects (\emph{where is the goal?}) and the local scene structure (\emph{how to get there?}). One classical approach would be to estimate the object's pose, from which the desired robot pose can be derived. But this usually requires specific object information such as 3D models \cite{wen2023foundationpose}, and the object being initially visible. This limits its applicability when the object 3D model is not available or the object is initially out of view. On the other hand, current learning-based navigation \cite{vint,spoc,saynav,gnm,feature-fields,nav-llm,mopa} is not quite applicable to real-world high-precision navigation tasks, as they lack precise goal conditioning, often assume discrete action spaces \cite{spoc, majumdar2022zson}, or trained on imprecise real-world data \cite{vint,gnm}.

In this work, we propose \emph{Aim My Robot} (\ours{}), an end-to-end vision-based local navigation model that navigates to objects with \emph{centimeter-level} precision. \ours{} does not require an object CAD model, and instead uses a reference image with object mask and relative pose for precise goal specification. \ours{} takes streams of RGB-D and LiDAR data as inputs and outputs trajectories directly, eliminating the need of a metric map. We achieve high precision and strong generalization via two contributions: 1) a data pipeline for generating large-scale, photorealistic, and precise trajectories to diverse objects; 2) a transformer-based model that takes multi-modal sensor inputs (RGB-D + LiDAR) and plan precise and safe trajectories. Experiments show that \ours{} achieves a median error of $3~\text{cm}$ and $1^\circ$ on unseen objects, with little degradation when deployed on a real robot. It supports robots of different sizes, and can be finetuned quickly to support more complex kinematics such as Ackermann steering vehicles. \ours{} is the first system for high-precision visual navigation with strong sim2real transfer. We hope it paves the way towards precision robot autonomy.

\begin{figure*}[t]
\centering
\includegraphics[width=1\textwidth]{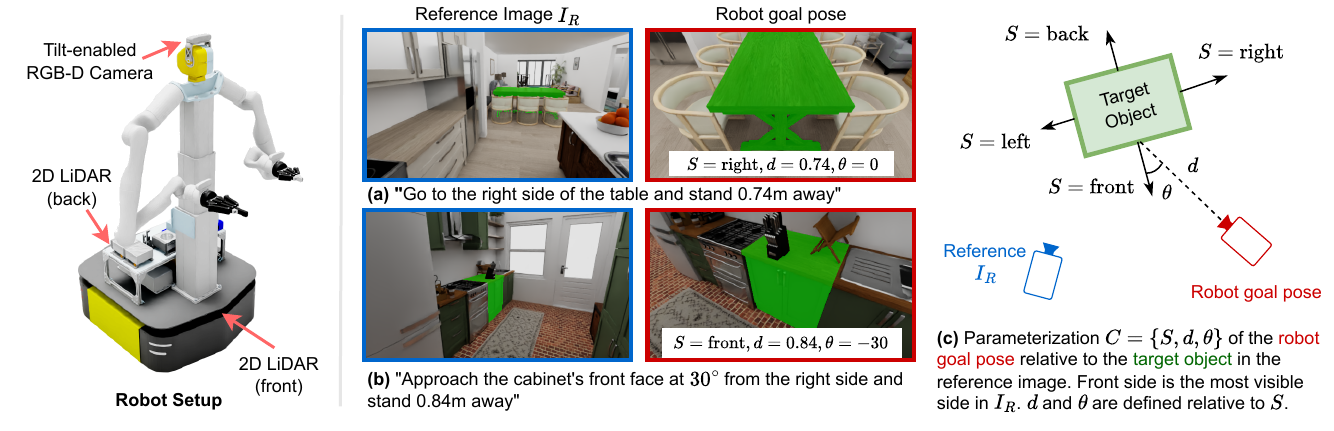}
\caption{\textbf{Problem setup.} We specify the target object via a reference image $I_R$ taken in the scene and an object mask $M$ (in green). The goal condition $\mathbf{C}$ is defined as the relative side and pose of the object in $I_R$. A robot needs to navigate to the object conditioned on $\mathbf{C}$ and tilt its camera to gaze at the object. Note the reference image does not represent the final image captured by the robot at the desired goal.}
\figurespace
\label{fig:task_spec}
\end{figure*}

\section{Problem Definition}
\label{sec:task}

We consider the scenario where the robot has reached the vicinity ($<$10\unit{\meter}, about a room size) of the object of interest and needs to perform a \emph{local}, \emph{object-centric} high-precision maneuver for docking, inspection and manipulation. We formulate this as a local navigation task where the goal $G$ consists of the target object and the relative base pose (in $SE(2)$) to the object. For generality, we assume $G$ is provided by another system. The robot has a tilt-enabled forward RGB-D camera at 1.5m high and a 2D LiDAR mounted on the base, providing 360$^{\circ}$ coverage (Fig.~\ref{fig:task_spec}). Concretely, the robot is given past sensor observations $\{o_t, o_{t-1}, ...\}$ along with the goal $G$, and needs to move its base to reach the specified pose while tilting its camera to gaze at the object. The robot does not have the object's 3D model or a 2D/3D map.

\textbf{Goal specification.} Due to the lack of the object's 3D model and a map, one challenge is how to unambiguously define the goal $G$. To address this, we assume the robot is given a reference image $I_R$ with the target object mask $M$ (Fig.~\ref{fig:task_spec}a and \ref{fig:task_spec}b). $I_R$ can be taken from the robot's long-term memory or from its recent observations. To make goal specification object-centric, we use the fact that common objects have 4 dominant sides (i.e. described by its bounding box). Denoting the most visible side in $I_R$ as the \emph{front}, then the relative pose can be defined as $\mathbf{C}=\{S, d, \theta\}$, where $S\in \{\text{front}, \text{back}, \text{left}, \text{right} \}$ is the approach side, $d\in [0.0\unit{\meter}, 1.0\unit{\meter}]$ is the approach distance, and $\theta\in \{0^\circ, \pm 15^\circ, \pm 30^\circ\}$ is the approach angle. See Fig.\ref{fig:task_spec}c for an illustration. Hence, $G=\{I_R, M, \mathbf{C}\}$.

\textbf{Discussion.} Existing object instance navigation systems require taking a close-up view for each object \cite{krantz2022instance} or mapping object locations \cite{chang2023goat,liu2024ok,bajracharya2024demonstrating}. In comparison, our formulation uses \emph{mask} to specify the object instance in an image. It does not require a close-up view or building a map. Moreover, it can be interfaced with a high-level task planner (e.g. SAM~\cite{sam}, TAMP \cite{garrett2020pddlstream} and LLM \cite{liang2023code}) that outputs mask and pose parameters whereby AMR guarantees \emph{precise} base and camera positioning.

\section{Aim My Robot}
In this section, we detail our technical approach. We first describe our data pipeline that generates large-scale, high-quality demonstrations entirely in simulation. Then, we introduce \ours{}, a multi-modal architecture for robust and precise local navigation.

\subsection{Data Generation}
\label{sec:data-gen}

Achieving strong generalization and high precision requires training data containing diverse scenes, objects, and precise trajectories. Since humans are poor at estimating distances and angles from images, we forgo teleoperation and leverage simulation and model-based planning for data collection.

\begin{figure}[t]
\centering
\includegraphics[width=1.0\columnwidth]{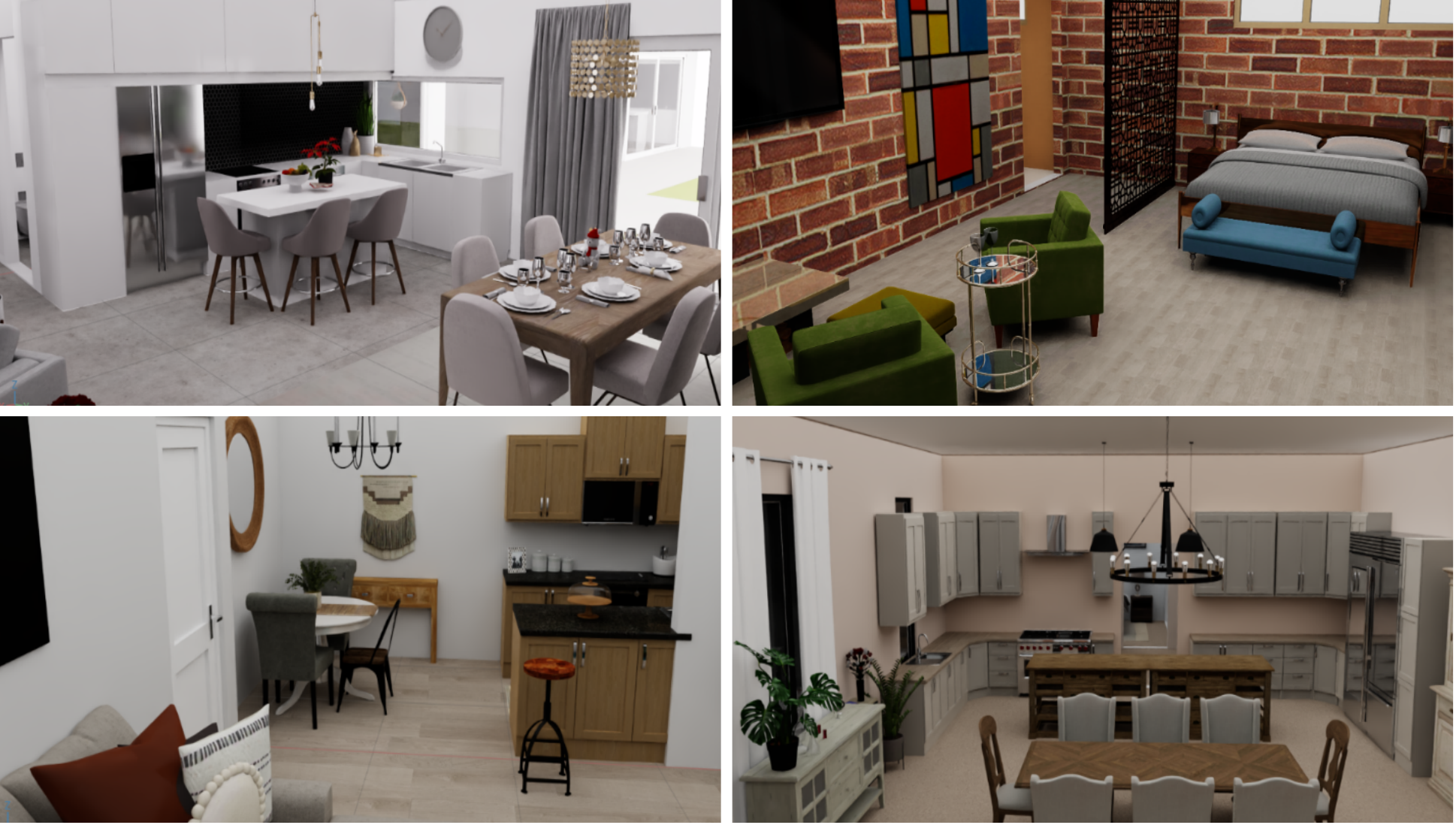}
\caption{Example rendered images of HSSD scenes in Isaac Sim.}
\label{fig:isaac-sim}
\end{figure}

\textbf{Assets and simulator.} We import the Habitat Synthetic Scenes Dataset (HSSD)~\cite{hssd} which contains diverse room layouts (100+) and objects (10,000+) into Isaac Sim. Since HSSD contains Physics-Based Rendering (PBR) textures, Isaac Sim is able to render photorealistic images using ray tracing (Fig.~\ref{fig:isaac-sim}). The diversity and realism of the perception data enable strong visual sim2real transfer.

\textbf{Sampling goal condition.} We model the robot as a cylindrical rigid body with a radius $R$ (i.e. its footprint) sampled from $[0.1\unit{\meter}, 0.5\unit{\meter}]$. The robot is randomly placed in the traversable area of a room. A reference image $I_R$ is rendered either from the robot's initial camera view or from a random camera view in the room. Then, the target object mask $M$ (obtained from the simulator) is randomly chosen from $I_R$. Non-objects such as walls and floors are excluded. We sample a goal specification $\mathbf{C}$ by randomly choosing a side $S\in\{\text{front}, \text{back}, \text{left}, \text{right}\}$, distance $d\in[0.1\unit{\meter}, 0.5\unit{\meter}]$ and angle $\theta\in \{0^\circ, \pm 15^\circ, \pm 30^\circ\}$. Finally, we place the robot at the goal and check for collisions.

\textbf{Trajectory generation.} The robot kinematic model is assumed to be a differential-drive robot. We run AIT*~\cite{ait} using the ReedsShepp state space with a turning radius of 0 for planning the base motion. It is straightforward to change the state space to model robots with other kinematics. The cost function encourages the camera to look toward the goal while penalizing excessive backward motion (small backward motion is allowed if the robot can reach the goal faster). For each feasible path, we render the camera observations along the path with a distance gap of $0.2\unit{\meter}$ or an angular gap of $5^\circ$. The camera tilt angle is set such that the lowest vertex of the object mesh appears at ¼ above the bottom of the image (even when the object is out of view). 

\begin{figure}[t]
\centering
\includegraphics[width=1.0\columnwidth]{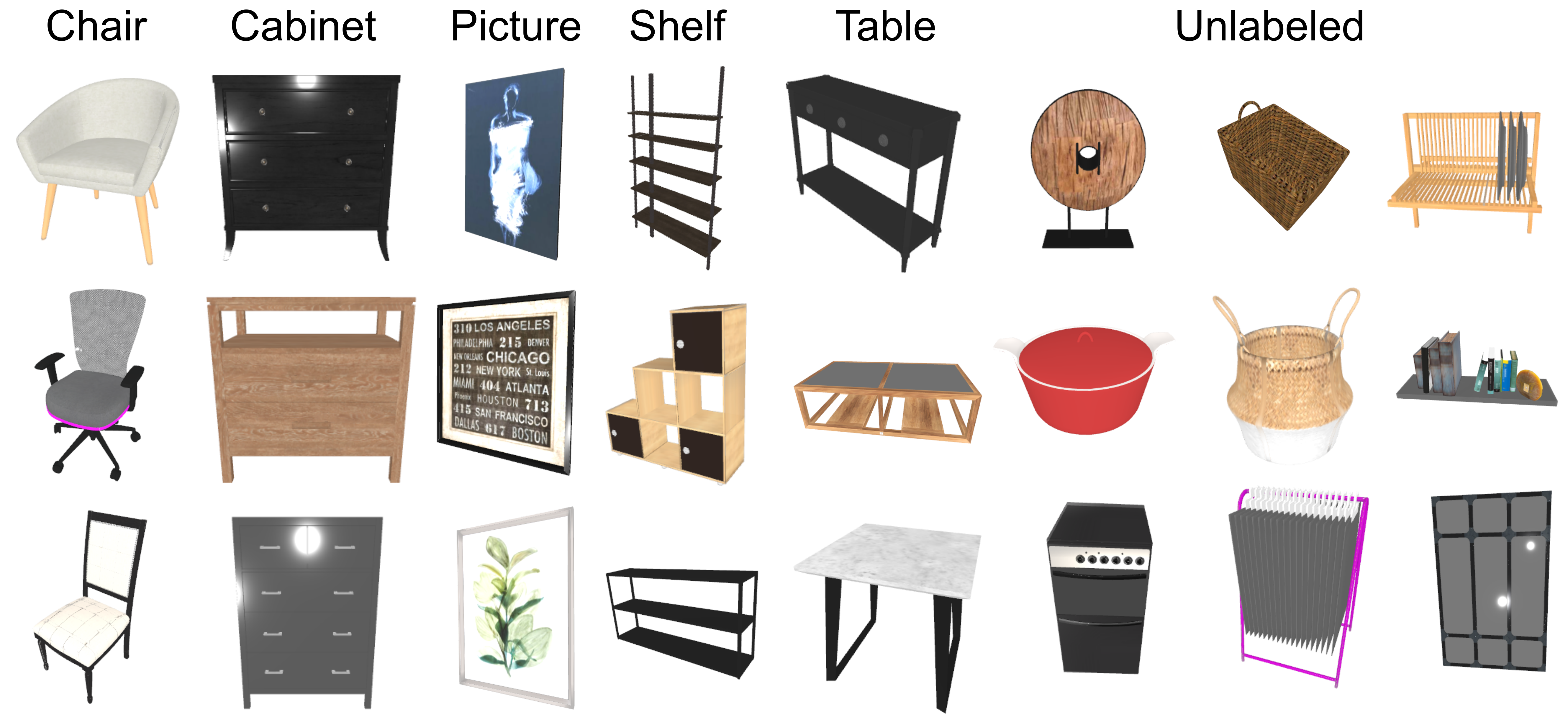}
\caption{\textbf{Sample objects in the scenes.} We consider all  objects, including those that are not semantically labeled.}
\label{fig:objects}
\figurespace
\end{figure}

\subsection{Model}
\label{sec:model}
\begin{figure*}[t]
\centering
\includegraphics[width=0.85\textwidth]{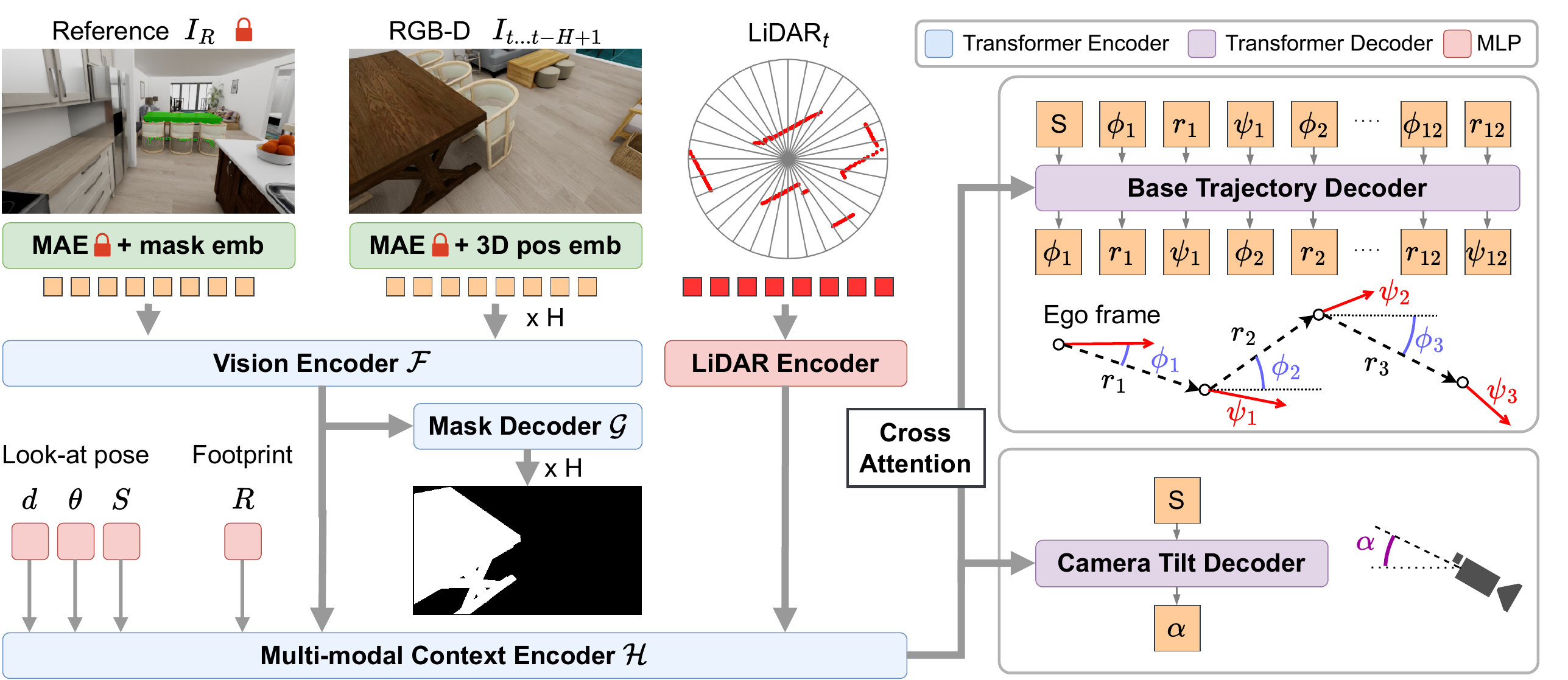}
\caption{\textbf{Network architecture.} The reference image $I_R$ and the robot's RGB-D observations $I_t$ are tokenized with MAE. The current LiDAR scan is tokenized by grouping points into directional bins. Image and LiDAR tokens are input into the multi-modal context encoder jointly with the look-at pose and footprint tokens. Finally, the output tokens of the context encoder are cross-attended to the base trajectory decoder and the camera tilt decoder. {\scriptsize \fbox{S}} are the learned start tokens for the decoders.}
\figurespace
\label{fig:model}
\end{figure*}

\ours{} is a transformer-based model (Fig.~\ref{fig:model}) based on two design principles: 1) a unified approach to representing multi-modal sensory data, and 2) an action decoding scheme that generates precise and collision-free actions. The system is divided into three stages: multi-modal sensor encoding, goal-aware sensor fusion, and multi-modal motion generation. We detail each stage as follows. 

\subsubsection{Encoding Multi-modal Sensor Data}
We use RGB, depth, and 2D LiDAR observations to achieve high precision and robustness. RGB is used for visual reasoning, depth provides precise 3D geometric information, and LiDAR ensures a $360^\circ$ coverage for robust collision avoidance. 

\textbf{Encoding RGB-D:} The current RGB image $I_t$ ($224\times 224\times 3$) is passed into a frozen Masked Autoencoder (MAE-Base) \cite{mae} to obtain a $14\times 14\times 512$ feature map. Each depth image is resized to $14\times 14$ and we compute the spatial location of each depth pixel in the robot's egocentric coordinate frame via $[x', y', z']=\mathbf{R}\mathbf{K}^{-1}[x, y, d]^T + \mathbf{t}$, where $\mathbf{K}$ is the camera intrinsics and $[\mathbf{R} | \mathbf{t}]$ is the camera extrinsics. We apply a sinusoidal position embedding $f$ for $x', y', z'$, respectively, and concatenate them to obtain the position embedding for each depth patch as $[f(x'); f(y'); f(z')]$. We add depth position embedding to each corresponding RGB patch. To incorporate past observations $I_{t-i}$, we set the camera extrinsics $[\mathbf{R}|\mathbf{t}]$ to be the transform between the camera pose at $t-i$ and the robot base pose at time $t$, which can be obtained from the robot's odometry. Compared to concatenating RGB and depth tokens which require twice number of tokens, using depth as positional encoding is more efficient since it keeps the number of visual tokens unchanged.

\textbf{Encoding LiDAR:} We resample the LiDAR points into 256 points. The points are grouped into 32 directional bins. The points in each bin (8 of $(x,y)$ coordinates) are passed into a Multi-layer Perceptron (MLP) to obtain one LiDAR token. In total, there are 32 LiDAR tokens.

\subsubsection{Goal-aware, Robot-aware Sensor Fusion}
We tokenize the reference image with the same frozen MAE. The mask $M$ is encoded by a shallow convolutional network, similar to \cite{sam}. All the visual tokens are flattened and passed to the Vision Context Encoder $\mathcal{F}$. The output tokens corresponding to the robot's observations are decoded into target masks using the mask decoder $\mathcal{G}$. This helps $\mathcal{F}$ to learn to track targets. We use separate MLPs to tokenize the goal $d, \theta, S$ and the robot footprint (i.e. radius) $R$. Finally, all the tokens are input to the Multi-modal Context Encoder $\mathcal{H}$. The output serves as the context for motion generation.

\subsubsection{Motion Generation}
The output of $\mathcal{H}$ is cross-attended to separate transformer decoders for the base movement and the camera tilt angle, respectively. \ours{} predicts base trajectory and camera tilt angle at different frequencies. For base trajectory, it operates in a receding-horizon fashion: the robot follows the predicted trajectory up to $T$ steps and then predicts a new trajectory given the updated robot observations. For camera tilt angle, it predicts a new value at every time step. %

\textbf{Base trajectory decoding:} The base trajectory is parameterized by a sequence of waypoints in egocentric polar coordinates. To capture the multi-modal nature of robot trajectories, we use an autoregressive transformer decoder. Since autoregressive classification requires discretizing the actions, to preserve the precision, we adopt multi-token classification with residual predictions. 
Specifically, we represent each waypoint by a sub-action tuple (direction $\psi_i\in[0, 2\pi]$, distance $r_i\in[0,0.2\text{m}]$, heading $\phi_i\in[0,2\pi]$), $i=0,...,T-1$ and sample $\psi, r, \phi$ conditionally in sequence. This representation reduces the number of bins required for classification, as classifying $\psi,r,\phi$ at once would require a combinatorial number of bins.  In practice, we discretize $\psi$, $r$, $\phi$ into 30, 32, 12 bins, respectively. For each output token $z$, we recover the continuous value $z'=C(z) + R(z, C(z))$ where $C(\cdot)$ is the output from the classifier, and $R(\cdot, \cdot)$ is an MLP that predicts the residual. %

\textbf{Camera tilt angle decoding:} Since camera tilt control is uni-modal, we continuously predict the camera tilt angle $\alpha$ via regression at each time step $t$.

\section{Implementation Details}
\label{sec:implementation}

\textbf{Dataset.} We converted 54 HSSD~\cite{hssd} environments into Universal Scene Description (USD) format and generated 500k trajectories (7.5~M frames) in Isaac Sim~\cite{isaacsim}. We use 49 scenes for training and 5 scenes for evaluation. There are 3,119 target objects in the training scenes and 275 target objects in the test scenes. Among the 275 test objects, 160 are unseen during training. We randomly sampled robot starting locations and target objects, and created 2000 navigation tasks in unseen environments for evaluation.

\textbf{Loss.}
We train the model with a supervised loss function:
$$L = L_\text{mask} + L_\text{base} + L_\text{tilt}$$
where $L_\text{mask}$ is pixel-wise L2 loss to regress the object mask in all history frames, $L_\text{base}$ is a sum of classification loss and regression loss for each waypoint, akin to \cite{behavior_transformer}, and $L_\text{tilt}$ is an L2 loss that regresses the camera tilt angle.

\textbf{Training.} We train \ours{} for 150k steps with a batch size of 128 on 8 A100 GPUs. After the initial Behavior Cloning phase, we ran the model in simulation and identified failures (collision, tracking loss, out of tolerance), and used DAgger~\cite{dagger} to augment the dataset. For experiments with HSSD, we trained all models without history ($\emph{hist=0}$), as we found that history did not show noticeable benefits in simulation. In our real robot experiment, we trained a model with 4 past frames and compare it against $\emph{hist=0}$ model.

\textbf{Deployment.} We deploy \ours{} on a real mobile manipulator with an omnidirectional mobile base and a simulated forklift with Ackermann steering in Isaac Sim. Each trajectory contains $T=12$ waypoints, separated by $dt=0.2s$. To track the predicted trajectories, we use a Pure Pursuit controller \cite{conlter1992implementation} for the omnidirectional robot and a Model Predictive Control (MPC) controller for the forklift. The camera tilt angle is updated at every time step, whereas the trajectory is updated with a new prediction at every 8 steps. The 
 models run at 12~Hz on an RTX 3090.

\section{Experiments}
\label{sec:result}

We perform \textbf{quantitative experiments} both in simulation and in the real world. Each run performs closed-loop inference until the robot comes to a stop, collides with obstacles, or exceeds the time limit. We consider a run \emph{complete} if the robot is within $1.0\unit{\meter}$ to the goal. Otherwise, we consider the run incomplete. The error distribution contains final pose error for only the completed runs, and the \emph{completion rate} is reported as a separate metric. In the real-world experiments, we measure the ground truth goal pose against an occupancy map with $2~\text{cm}$ resolution. The robot's ground truth pose is obtained by classical Monte Carlo Localization~\cite{mclocalization}.

We conduct additional \textbf{qualitative experiments} to highlight generalization to novel tasks using \ours{}: We show generalization to new tasks and scenes in the real-world, closing a fridge drawer using the robot body, and a forklift picking up a pallet.

\subsection{Simulation Results with HSSD}

\textbf{Baseline:} We compare our method against a classical pipeline based on object pose estimation. We use FoundationPose \cite{wen2023foundationpose} to estimate the pose of the target object in the initial observation given the CAD model. The posed bounding box is used to compute the goal pose as described in Sec.~\ref{sec:task}. Then, we run the same planner to find a path using a groundtruth occupancy map, and navigate the robot to the goal. Note that such a system uses extra information not required by \ours{}: 1) a CAD model of the target object; 2) the object is visible in the robot's initial observation (required by pose estimation); and 3) perfect mapping.

\begin{figure}[t]
\centering
\includegraphics[width=0.95\columnwidth]{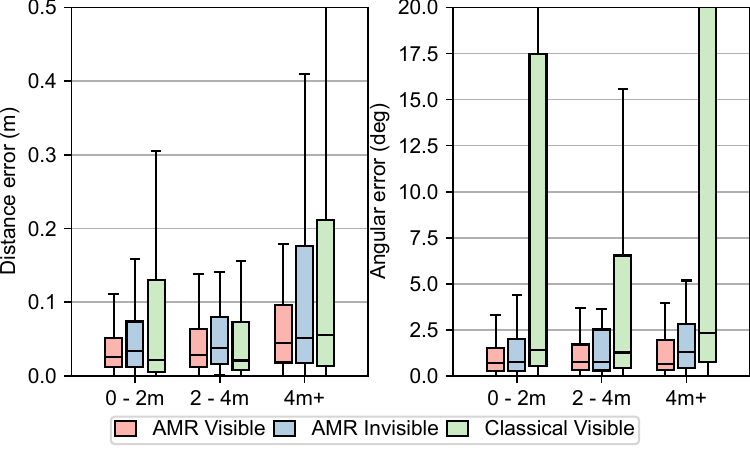}
\footnotesize
\setlength{\tabcolsep}{4pt}

\begin{tabular}[t]{cccc}
& AMR Visible & AMR Invisible & Classical Visible\\
\hline
Completion\% & 95.9 & 90.4 & 89.0
\end{tabular}

\caption{\textbf{Navigation error distribution in the test scenes for various object distances for completed runs. \ours{} outperforms the classical baseline for when objects are initially visible; and \ours{} also outperforms classical pose estimation when the object is initially out-of-view.} 
Our approach achieves higher completion rate than the classical baseline method overall. Note, classical pose estimation requires the object to be initially visible. }
\label{fig:baseline-comparison}
\end{figure}

\textbf{Quantitative Results:} Fig.~\ref{fig:baseline-comparison} compares the error distribution between \ours{} and the classical approach. We consider two cases: \emph{visible}, where the initial observation is used as the reference image, and \emph{invisible}, where the object is initially out-of-view. The classical approach can only be used in the \emph{visible} case.  
In the \emph{visible} cases, \ours{} performs better than the baseline approach even when the object is far. 
In the \emph{invisible} cases, our approach outperforms the baseline approach with similar accuracy and significantly lower variance. One interesting observation is that the angular error for the baseline increases when objects are too close ($0-2\unit{\meter}$). This is usually caused by large objects (e.g. furniture) being partially out of view. In contrast, \ours{} actively moves the robot, making it more robust to state estimation errors.

In Table~\ref{tab:errors} we compare the median pose errors between the seen objects and unseen objects in the test scenes. We do not see a clear gap, showing that \ours{} generalizes well to unseen objects.

\begin{table}[t]
\centering
\begin{tabularx}{\columnwidth}{lYYYY}
\toprule
 & \multicolumn{2}{c}{\textbf{Seen} (136 unique objects)} & \multicolumn{2}{c}{\textbf{Unseen} (166 unique objects)} \\
Category & MAE~($^\circ$) & MDE~(\unit{\meter}) & MAE~($^\circ$) & MDE~(\unit{\meter})\\
\midrule
Chair & 1.45 & 0.05 & 1.08 & 0.04 \\
Drawers & 0.77 & 0.02 & 0.55 & 0.03 \\
Couch & 0.82 & 0.02 & 0.57 & 0.04 \\
Picture & 0.84 & 0.03 & 0.64 & 0.04 \\
Shelves & 0.65 & 0.07 & 0.68 & 0.02 \\
Tables & 7.14 & 0.04 & 1.20 & 0.04 \\
Unlabeled & 0.78 & 0.03 & 0.74 & 0.04 \\
\bottomrule
\end{tabularx}
\caption{Comparison of median angular errors (MAE) and median distance errors (MDE) for seen and unseen instances across shared categories in the test scenes.}
\label{tab:errors}
\figurespace
\end{table}

\begin{figure*}[t]
    \centering
    \includegraphics[width=1\textwidth]{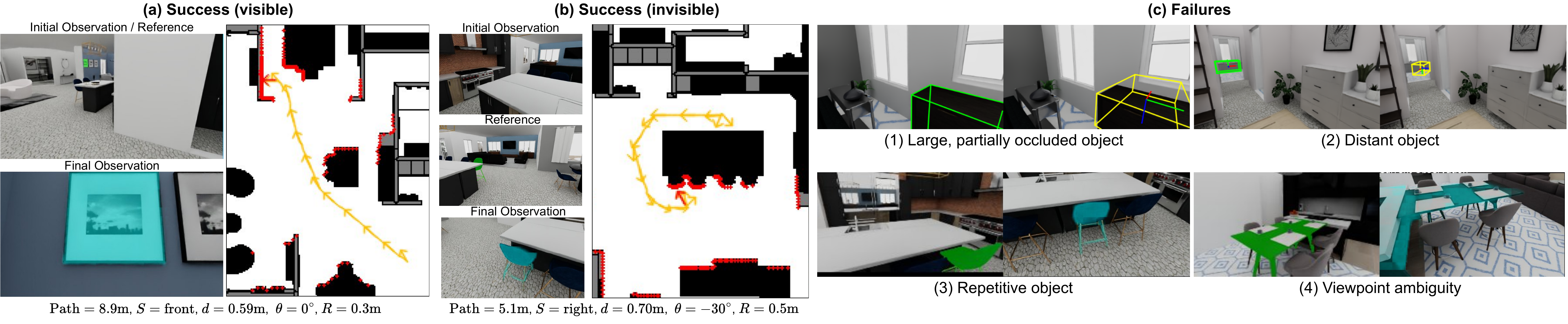}
\caption{\textbf{Qualitative examples:} \textbf{(a)} object initially visible \textbf{(b)} object initially out of view. The initial masks (green) are specified in the reference images . Cyan masks are predicted by the model. Trajectory accommodates the robot radius $R$ for obstacle avoidance. \textbf{(c) Typical failures:} (1,2) Baseline: Object pose estimation may fail when the object is partially occluded or too far (green box: groundtruth, yellow box: prediction) (3) \ours{} may go to the wrong object when the object is repetitive. (4) AMR and Baseline: a robot may go to the wrong side when there is no dominantly visible side (i.e. looking at an object from $45^\circ$ angle). This can be addressed by using a less ambiguous reference image. }
\label{fig:qualitative}
\figurespace
\end{figure*}

\textbf{Qualitative results:} Fig.\ref{fig:qualitative}a and \ref{fig:qualitative}b present two successful runs with different goal poses and robot sizes. \ours{} plans safe trajectories by considering the robot's size, and is able to track and reach target object precisely even when the target is far or out of view. Fig.\ref{fig:qualitative}c shows typical failure cases for both the baseline and \ours{}. For the baseline, the failure is mostly caused by pose estimation in challenging scenarios, such as occlusion and object being too far. \ours{} sometimes fails to track the correct object when there are repetitive objects, and may go to the wrong side when the viewpoint is ambiguous.

\subsection{Ablation Study} \label{sec:ablation}
\begin{figure}[t]
\centering
\includegraphics[width=0.95\columnwidth]{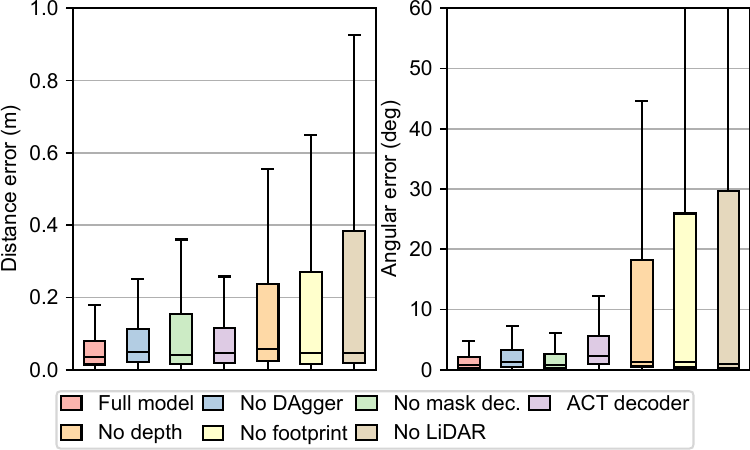}
\caption{Ablation study demonstrating effects of various choices on navigation errors in the test scenes. We consider both complete and incomplete runs.}
\label{fig:ablation}
\figurespace
\end{figure}

To verify our design choices, we ablate our models by disabling some of the components and analyze their impacts on the precision (Fig.~\ref{fig:ablation}) and robustness (Table~\ref{tab:collision-rate}). In particular, we find \textbf{Sensor Modality} has the biggest impact. Without LiDAR, the robot is more prone to colliding with surrounding objects. Likewise, the 3D information from depth helps the robot avoid obstacles and approach targets accurately. Without the \textbf{Footprint} token the model suffers from a high collision rate, because the robot cannot take its size into consideration when planning. Our \textbf{Autoregressive Trajectory Decoding} scheme outperforms Action Chunking Transformer (ACT) since it better captures the multi-modal nature of navigation trajectories. While \textbf{Mask Tracking} only moderately improves the precision, it improves the model interpretability, as we find strong correlation between incorrect mask tracking and robot navigating to the wrong object.

\begin{table}[h]
  \centering
  \resizebox{\columnwidth}{!}{
  \setlength{\tabcolsep}{4pt}
  \begin{tabular}{cccccccc}
  \toprule
    Full & No mask dec. & ACT decoder & No DAgger & No depth & No footprint & No LiDAR\\
    \midrule
    \textbf{3.8} & 7.1 & 5.8 & 5.5 & 17.7 & 26.2 & 25.6\\
    \bottomrule
  \end{tabular}
  }
  \caption{Collision rate of ablated models.}
  \label{tab:collision-rate}
  \figurespace
\end{table}

\textbf{Large-scale evaluation matters:} In Fig.~\ref{fig:ablation}, all ablated models have comparable median errors, indicating that they perform almost equally well for more than half of the test cases, but their long-tail failure cases vary significantly. This implies that our method is more \emph{robust} across large datasets with environment variations.

\subsection{Closed-loop Hardware Evaluation}

\begin{figure*}[t]
\centering
\includegraphics[width=0.8\textwidth]{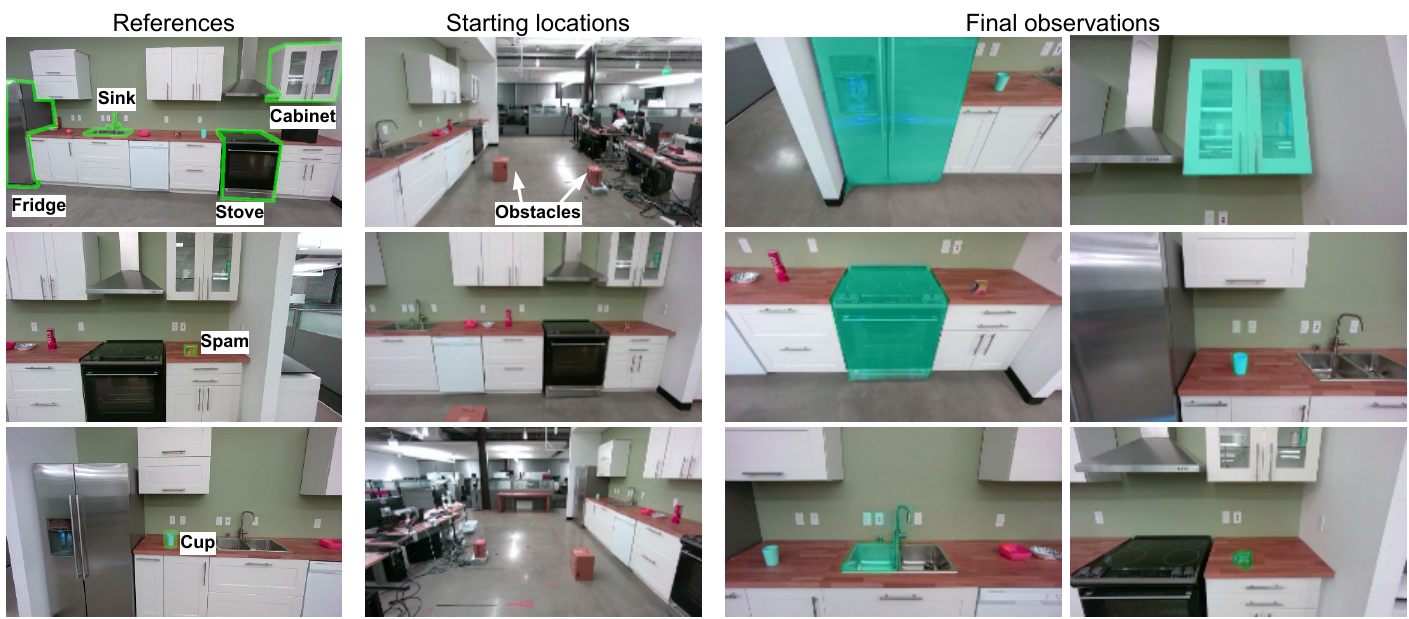}
\caption{\textbf{Real kitchen experiments.} \textbf{Left:} reference images and contours of target object masks. While the \emph{fridge} is partially out of view, the model can still reach the fridge. \textbf{Middle:} Initial view of the scene through the tilt camera. \textbf{Right:} robot observations after reaching each object overlaid with predicted masks (in cyan).}
\label{fig:real-kithcen}
\end{figure*}

We evaluate our system in a real kitchen on a mobile manipulator with an omnidirectional base and an RGB-D head camera with tilt support. We use one reference image covering the whole kitchen to specify 4 large objects (\emph{fridge, sink, stove} and \emph{cabinet}) and two zoomed-in images for small targets (\emph{spam} and \emph{cup}). The robot is initialized at three distinct starting locations. For each run, we continuously run \ours{} to navigate the robot to all 6 objects, with the relative goal condition describing the robot facing the object of interest, i.e., $\mathbf{C}$=$(d=1~\text{m}, \theta=0^\circ, S=\text{front})$. For each object, we perform 3 runs.

In Table~\ref{tab:real-kitchen}, we show the number of completed trials and compute the errors against hand-measured ground truths. $\emph{hist=4}$ succeeded in all tasks except for \emph{spam} starting at location \#3. Unlike in simulation, history is critical for the robot to track the target, as $\emph{hist=0}$ fails more often on large (\emph{fridge}) and small (\emph{spam}) objects. Among the objects, $\emph{sink}$ has the largest distance error due to unstable tracking and the lack of a large surface area, as the sink is an object that is concave to the tabletop surface and can easily be obscured. With the exception of \emph{sink} and \emph{cup}, all other objects achieve distance errors on the order of $1.8 \sim 3.1$ cm to the goal specification.

\begin{table}[t]
  \footnotesize
  \begin{subtable}{1\columnwidth}
  \centering
  \setlength{\tabcolsep}{2.5pt}
  \begin{tabular}{C{1cm}C{2.2cm}C{2.2cm}C{2.2cm}}
    \toprule
      & Fridge & Sink & Stove \\
     \midrule
     Hist=0 
     & 1 / $2.4~\text{cm}$ / $2.3^\circ$ 
     & \cellcolor{green!25} 3 / $14.8~\text{cm}$ / $0.1^\circ$ 
     & \cellcolor{green!25} 3 / $2.0~\text{cm}$ / $0.8^\circ$ \\
     Hist=4 
     & \cellcolor{green!25} 3 / $3.1~\text{cm}$ / $0.4^\circ$ 
     & \cellcolor{green!25} 3 / $7.9~\text{cm}$ / $0.2^\circ$ 
     & \cellcolor{green!25} 3 / $2.4~\text{cm}$ / $0.8^\circ$ \\
  \end{tabular}
  \end{subtable}
  \bigskip
  \begin{subtable}{1\columnwidth}
  \centering
  \setlength{\tabcolsep}{2.5pt}
  \begin{tabular}{C{1cm}C{2.2cm}C{2.2cm}C{2.2cm}}
    \toprule
     & Cabinet & Spam & Cup \\
     \midrule
     Hist=0 
     & \cellcolor{green!10} 2 / $3.1~\text{cm}$ / $1.7^\circ$ 
     &  1 / $1.8~\text{cm}$ / $0.6^\circ$ 
     & \cellcolor{green!25} 3 / $9.6~\text{cm}$ / $0.8^\circ$\\
     Hist=4 
     & \cellcolor{green!25} 3 / $2.1~\text{cm}$ / $0.6^\circ$ 
     & \cellcolor{green!10} 2 / $2.0~\text{cm}$ / $0.1^\circ$ 
     & \cellcolor{green!25} 3 / $8.7~\text{cm}$ / $0.9^\circ$\\
    \bottomrule
  \end{tabular}
  \end{subtable}
  \caption{\textbf{Mean navigation errors in a real kitchen.} For each object we report (\#completed / distance error / orientation error). 3 runs were performed per object. Cells are colored green according to the number of completed runs.}
  \label{tab:real-kitchen}
  \figurespace
\end{table}

\subsection{Qualitative Experiments}

\begin{figure*}[t]
\centering
    \includegraphics[width=0.8\textwidth]{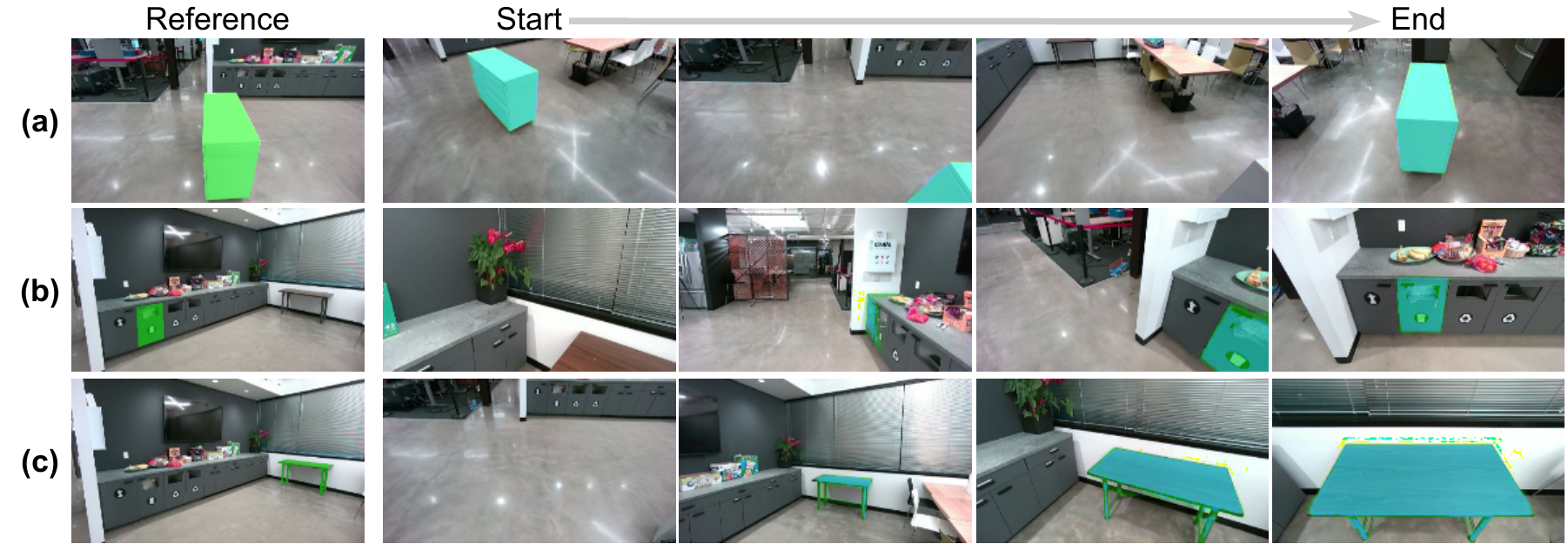}
\caption{\textbf{Real-world qualitative experiments in new scenes.} \textbf{Left:} reference image with target object highlighted by the green mask. \textbf{Remaining columns:} robot's camera view while moving. The tracked object is highlighted by the cyan mask. The tasks are: \textbf{(a)} Go to the back of the cabinet and stand 0.8\unit{\meter} away. \textbf{(b)} Go to the front of the landfill trashcan and stand 1\unit{\meter} away. \textbf{(c)} Go to the front of the table and stand 0.5\unit{\meter} away.}
\label{fig:real-qualitative}
\end{figure*}

\begin{figure*}[t]
\centering
\includegraphics[width=1\textwidth]{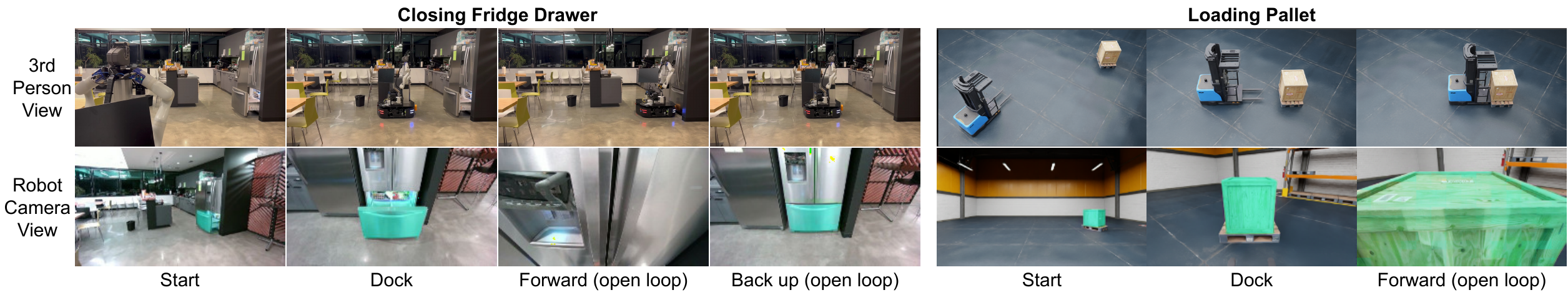}
\caption{\textbf{Using \ours{} for mobile manipulation tasks.} \textbf{Left:} closing a fridge drawer by pushing the robot body towards the drawer. \textbf{Right:} forklift loading a pallet. \ours{} accurately tracks the target objects throughout.}
\label{fig:close-fridge-drawer}
\end{figure*}

\textbf{Generalization in new real-world scenes:} In Fig.~\ref{fig:real-qualitative}, we show additional results of \ours{} navigating to diverse objects in another real-world environment. \ours{} is able to handle objects being out of view and large viewpoint change (i.e. going to the back of a cabinet). We refer the reader to the website for the videos.

\textbf{Closing the fridge drawer:} In Fig.\ref{fig:close-fridge-drawer}, we show that by accurately aiming the robot to a target object a robot can perform downstream tasks using its body as the end effector. The goal is specified by $M=I_\text{drawer}$ and $\mathbf{C}=(d=1~\text{m}, \theta=0^\circ, S=\text{front})$. Once the robot reaches the goal, the robot moves forward to close the drawer in open loop. 

\textbf{Loading a pallet in a warehouse:} We test \ours{} on a simulated forklift to load a randomly placed pallet in a warehouse. Existing work on controlling forklifts requires sophisticated modeling and motion planning \cite{tamba2009path, sun2023optimization} with privileged state information. Here, we show that a learning system with onboard sensors can achieve the same precision. We empirically generate $\sim500$ demonstrations with forklift kinematics to finetune \ours{}. In Fig.\ref{fig:close-fridge-drawer}, we show \ours{} successfully navigates a forklift to face a pallet directly $\mathbf{C}$=$(d$=$2~\text{m}, \theta$=$0^\circ, S$=$\text{front})$. The precision is sufficient such that the fork can be completely inserted into a pallet by driving the forklift forward in an open loop fashion. More information is provided on the website.

\section{Related Work}
\label{sec:related-work}

\textbf{Object-goal and instance-goal navigation:}
Object-goal navigation typically refers to a robot navigating to any object of the specified category (e.g., couch) \cite{gervet2023navigating, majumdar2022zson, chaplot2020object}. Instance-goal navigation enables a robot to reach a specific object by using a close-up view of that object \cite{krantz2022instance, krantz2023navigating, chang2023goat}. These systems typically consider the goal reached when the robot is within $1.0\unit{\meter}$ radius to the goal. While \cite{wasserman2023last} improves the last-mile goal approaching through image matching, it is unable to reach a precise pose relative to an object. Unlike existing approaches, \ours{} is designed to reach any object with centimeter level precision. Such generality and precision are enabled by a novel goal parametrization (reference image with mask), and a model trained on diverse and precise demonstrations.

\textbf{Object pose estimation:} Object pose estimation \cite{wen2023foundationpose, sam6d} provides an alternative solution to high-precision object-centric navigation. State-of-the-art object pose estimation typically requires prior knowledge of the object (CAD model, template images, or object category), and the object must be visible in the camera. These assumptions are difficult to satisfy since a robot often needs to go to arbitrary objects and the object can be temporarily out-of-view. In contrast, \ours{} is trained on diverse and photorealistic simulation data to learn general knowledge of object instances and geometries. Furthermore, unlike object pose estimation which is ``passive'' and struggles when the object is too small or too far, \ours{} is an active approach akin to Image-based Visual Servoing (IBVS) \cite{wang2024nerf} that refines its predictions as the robot gets closer to the  goal. \ours{} is more sophisticated than IBVS since \ours{} reasons about objects, adheres to robot kinematics and handles collision avoidance.

\textbf{Mobile manipulation systems:}
There is a surge of interest in developing learning-based mobile manipulation systems \cite{spoc,uppal2024spin,yokoyama2023asc,mobile-aloha,skilltransformer,hu2023causal}. Some works assume additional information such as object pose or robot pose to be available (\cite{uppal2024spin, yokoyama2023asc, hu2023causal, skilltransformer}). Other systems show that it is possible to directly map sensor observations to actions \cite{spoc,mobile-aloha}, but they only consider a limited set of scenarios (e.g. no need to avoid unseen obstacles or handle unseen objects), so that high-precision base positioning is not a major concern. As of today, the most general and capable mobile manipulation systems are still modular \cite{liu2024ok, bajracharya2024demonstrating}, and they require mapping the environment and objects beforehand. In particular, \cite{liu2024ok} shows that accurate base positioning is crucial for manipulation to succeed. Our work can be integrated into both learning and modular mobile manipulation systems so that a robot can reach an accurate base pose for manipulation without a prior map or object model.

\section{Conclusion and Limitations}
\label{sec:limitation}
We present \ours{}, a vision-based navigation model that navigates to any object with centimeter precision. While trained completely in simulation, it transfers to real-world and unseen objects with little degradation in precision. \ours{} does not require object 3D models or a map to operate, runs in real-time at 10~Hz, and easily adapts to other robots with fine-tuning.

\textbf{Limitations.} \ours{} is designed for local navigation as it is equipped with a short-term memory. Since it is only trained in household environments, its generalization in other scenarios such as factories and unstructured environments is potentially limited. Additionally, real robots have complex geometric shapes and sensor placements, and our assumption of a robot being cylindrical with a centered camera can hinder its applicability to diverse mobile robots such as legged robots. However, we anticipate that these limitations can be addressed by increasing the robot's memory window, improving the diversity of the training environments, and more comprehensive modeling of robot shape and sensor placements.

\addtolength{\textheight}{-1cm}

\bibliographystyle{IEEEtran}
{\scriptsize
\bibliography{IEEEabrv,references.bib}}

\end{document}